\documentclass[a4paper]{article}

\usepackage{iwslt18,amssymb,amsmath,epsfig}
\def\reg{{\rm\ooalign{\hfil
     \raise.07ex\hbox{\scriptsize R}\hfil\crcr\mathhexbox20D}}}

\usepackage[utf8]{inputenc}
\usepackage{lipsum}
\usepackage{multirow}

\newcommand{\sep}{\thinspace|\thinspace}

\newcommand{\amun}{\nobreak\textsc{Amun}}

\newcommand{\xf}{\nobreak\textsc{Transformer}}

\newcommand{\tedlium}{\nobreak\textsc{TED-LIUM}}
\newcommand{\tedtrans}{\nobreak\textsc{TED-Trans}}
\newcommand{\opensubs}{\nobreak\textsc{OpenSubtitles2018}}

\newcommand{\subs}{\nobreak\textsc{Subs}}
\newcommand{\subsasr}{\nobreak\textsc{Subs-ASR}}
\newcommand{\subsasrlc}{\nobreak\textsc{Subs-ASR-lc}}

\newcommand{\tedasr}{\nobreak\textsc{TED-ASR}}
\newcommand{\tedasrtop}[1]{\nobreak\textsc{TED-ASR-Top-$#1$}}

\newcommand{\asrref}{\nobreak\textsc{ASR-ref}}
\newcommand{\asrout}{\nobreak\textsc{ASR-out}}

\title{The MeMAD Submission to the IWSLT 2018 Speech Translation Task}

\makeatletter
\def\name#1{\gdef\@name{#1\\}}
\makeatother

\name{{\em Umut Sulubacak\thinspace$^\ast$\hspace{1em}Jörg Tiedemann\thinspace$^\ast$\hspace{1em}Aku Rouhe\thinspace$^\dagger$\hspace{1em}Stig-Arne Grönroos\thinspace$^\dagger$\hspace{1em}Mikko Kurimo\thinspace$^\dagger$}}

\address{
    $\ast$~Department of Digital Humanities / HELDIG~~~~\\
    University of Helsinki, Finland \\
    {\small \tt \{umut.sulubacak{\sep}jorg.tiedemann\}@helsinki.fi} \\[1em]
    $\dagger$~Department of Signal Processing and Acoustics~~~~\\
    Aalto University, Finland \\
    {\small \tt \{aku.rouhe{\sep}stig-arne.gronroos{\sep}mikko.kurimo\}@aalto.fi}
}

\begin{document}
\maketitle


\begin{abstract}
This paper describes the MeMAD project entry to the IWSLT Speech Translation Shared Task, addressing the translation of English audio into German text. Between the pipeline and end-to-end model tracks, we participated only in the former, with three contrastive systems. We tried also the latter, but were not able to finish our end-to-end model in time.

All of our systems start by transcribing the audio into text through an automatic speech recognition~(ASR) model trained on the TED-LIUM English Speech Recognition Corpus~(\tedlium{}). Afterwards, we feed the transcripts into English-German text-based neural machine translation~(NMT) models. Our systems employ three different translation models trained on separate training sets compiled from the English-German part of the TED Speech Translation Corpus~(\tedtrans{}) and the \opensubs{} section of the OPUS collection.

In this paper, we also describe the experiments leading up to our final systems. Our experiments indicate that using \opensubs{} in training significantly improves translation performance. We also experimented with various pre- and postprocessing routines for the NMT module, but we did not have much success with these.


Our best-scoring system attains a BLEU score of~16.45 on the test set for this year's task.
\end{abstract}


\section{Introduction}
\label{sec:intro}

The evident challenge of speech translation is the transfer of implicit semantics between two different modalities. An end-to-end solution to this task must deal with the challenge posed by intermodality simultaneously with that of interlingual transfer. In a traditional pipeline approach, while speech-to-text transcription is abstracted from translation, there is then the additional risk of error transfer between the two stages. The MeMAD project\footnote{https://www.memad.eu/} aims at multilingual description and search in audiovisual data. For this reason, multimodal translation is of great interest to the project.

Our pipeline submission to this year's speech translation task incorporates one ASR model and three contrastive NMT models. For the ASR module, we trained a time-delay neural network (TDNN) acoustic model using the Kaldi toolkit~\cite{Povey_ASRU2011} on the provided \tedlium{} speech recognition corpus~\cite{tedlium}. We used the transformer implementation of MarianNMT~\cite{mariannmt} to train our NMT models. For these models, we used contrastive splits of data compiled from two different sources: The $n$-best decoding hypotheses of the \tedtrans{}~\cite{tedtrans} in-domain speech data, and a version of the \opensubs{}~\cite{opensubtitles} out-of-domain text data~(\subs{}), further ``translated'' to an ASR-like format~(\subsasr{}) using a sequence-to-sequence NMT model. The primary system in our submission uses the NMT model trained on the whole data including~\subsasr{}, whereas one of the two contrastive systems uses the original \subs{} before the conversion to an ASR-like format, and the other omits \opensubs{} altogether.

We provide further details about the ASR module in Section~\ref{sec:asr}. Later, we provide a review of our experiments on the NMT module in Section~\ref{sec:trans}. The first experiment we describe involves a pre-processing step where we convert our out-of-domain training data to an ASR-like format to avoid mismatch between source-side training samples. Afterwards, we report a postprocessing experiment where we retrain our NMT models with lowercased data, and defer case restoration to a subsequent procedure, and another where we translate several ASR hypotheses at once for each source sample, re-rank their output translations by a language model, and then choose the best-scoring translation for that sample. We present our results in Section~\ref{sec:results} along with the relevant discussions.


\section{Speech Recognition}
\label{sec:asr}
The first step in our pipeline is automatic speech recognition. The organizers provide a baseline ASR implementation, which consists of a single, end-to-end trained neural network using a Listen, Attend and Spell~(LAS) architecture~\cite{chan2016listen}. The baseline uses the XNMT toolkit~\cite{neubig2018xnmt}. However, we were not able to compile the baseline system, so we trained our own conventional, hybrid TDNN-HMM ASR system using the Kaldi toolkit.

\subsection{Architecture}
\label{sec:asr:architecture}
Our ASR system uses the standard Kaldi recipe for the \tedlium{} dataset (release 2), although we filter out some data from the training set to comply with the IWSLT restrictions. The recipe trains a TDNN acoustic model using the lattice-free maximum mutual information criterion~\cite{povey2016purely}. The audio transcripts and large amount of out-of-domain text data included with the TED-LIUM dataset are used to train a heavily pruned 4-gram language model for first-pass decoding and less pruned 4-gram model for rescoring.

\subsection{Word Error Rates}
\label{sec:asr:wer}
The LAS architecture has achieved state-of-the-art word error rates (WER) on a task with two orders of magnitude more training data than here~\cite{chiu2017state}, but on smaller datasets hybrid TDNN-HMM ASR approaches are still considerably better. Table~\ref{tab:asr:wer} shows the results of our ASR model contrasted with those reported by XNMT in~\cite{neubig2018xnmt}, on the \tedlium{} development and test sets.

\begin{table}[htbp]
    \begin{center}
    \begin{tabular}{lrr}
        \hline
        Model & Dev WER & Test WER  \\
        \hline
        TDNN $+$ large 4-gram  & 8.24 & 8.83 \\
        LAS & 15.83 & 16.16 \\
        \hline
    \end{tabular}
    \end{center}
    
    \caption{Word error rates on the \tedlium{} dataset.}
    \label{tab:asr:wer}
\end{table}


\section{Text-Based Translation}
\label{sec:trans}

The ASR stage of our pipeline effectively converts the task of speech translation to text-based machine translation. For this stage, we build a variety of NMT setups and assess their performances. We experiment variously with the training architecture, different compositions of the training data, and several pre- and postprocessing methods. We present these experiments in detail in the subsections to follow, and then discuss their results in Section~\ref{sec:results}.

\subsection{Data Preparation}
\label{sec:trans:data}

We used the development and test sets from 2010's shared task for validation during training, and the test sets from the tasks between 2013 and 2015 for testing performance during development. In all of our NMT models, we preprocessed our data using the punctuation normalization and tokenization utilities from Moses~\cite{moses}, and applied byte-pair encoding~\cite{sennrich2015neural} through full-cased and lowercased models as relevant, trained on the combined English and German texts in \tedtrans{} and \subs{} using 37,000 merge operations to create the vocabulary.

We experiment with attentional sequence-to-sequence models using the Nematus architecture~\cite{Sennrich2017} with tied embeddings, layer normalization, RNN dropout of 0.2 and source/target dropout of 0.1. Token embeddings have a dimensionality of 512 and the RNN layer units a size of 1024. The RNNs make use of GRUs in both, encoder and decoder. We use validation data and early stopping after five cycles (1,000 updates each) of decreasing cross-entropy scores. During training we apply dynamic mini-batch fitting with a workspace of 3GB. We also enable length normalization.

For the experiments with the transformer architecture we apply the standard setup with six layers in encoder and decoder, eight attention heads and a dynamic mini-batch fit to 8GB of work space. We also add recommended options such as transformer dropout of 0.1, label smoothing of 0.1, a learning rate of 0.0003, a learning-rate warmup with a linearly increasing rate during the first 16,000 steps, a decreasing learning rate starting at 16,000 steps, a gradient clip norm of 5 and exponential smoothing of parameters.

All translations are created with a beam decoder of size 12.

\subsubsection{ASR Output for TED Talks}
\label{sec:trans:data:ted}

Translation models trained on standard language are not a good fit for a pipeline architecture that needs to handle noisy output from the ASR component discussed previously in Section~\ref{sec:asr}. Therefore, we ran speech recognition on the entire \tedtrans{} corpus in order to replace the original, human-produced English transcriptions with ASR output, which has realistic recognition errors.

To generate additional speech recognition errors to the training transcripts, we selected the top-$50$ decoding hypotheses. We did the same also for the development data to test our approach.
We can now sample from those ASR hypotheses to create training data for our translation models that use the output of English ASR as its input. We experimented with various strategies varying from a selection of the top $n$ ASR candidates to different mixtures of hypotheses of different ranks of confidence. Some of these are shown in Table~\ref{tab:trans:data:ted}. In the end, there was not a lot of variance between the scores resulting from this selection, and we decided to use the top-$10$ ASR outputs in the remaining experiments to encourage some tolerance for speech recognition errors in the system.

\begin{table}[htbp]
    \begin{center}
    \begin{tabular}{llr}
    \hline
    Training data  & Model   & BLEU  \\
    \hline
    \tedasrtop{1}  & \amun{} & 16.65 \\
    \tedasrtop{10} & \amun{} & 16.28 \\
    \tedasrtop{50} & \amun{} & 15.88 \\
    \hline
    \tedasrtop{1}  & \xf{}   & 18.25 \\
    \tedasrtop{10} & \xf{}   & 17.90 \\
    \tedasrtop{50} & \xf{}   & 18.14 \\
    \hline
    \end{tabular}
    \end{center}
    
    \caption{Translating the development test set with different models and different selections of ASR output and German translations from the parallel \tedtrans{} training corpus.}
    \label{tab:trans:data:ted}
\end{table}

\subsubsection{Translating Written English to ASR-English}
\label{sec:trans:data:subs:top-n}

The training data that includes audio is very limited and much larger resources are available for text-only systems. Especially useful for the translation of TED talks is the collection of movie subtitles in \opensubs{}. For English-German, there is a huge amount of movie subtitles (roughly 22 million aligned sentences with over 170 million tokens per language) that can be used to boost the performance of the NMT module. 

The problem is, of course, that the subtitles come in regular language, and, again, we would see a mismatch between the training data and the ASR output in the speech translation pipeline. In contrast to approaches that try to normalize ASR output to reflect standard text-based MT input such as~\cite{matusov}, we had the idea to transform regular English into ASR-like English using a translation model trained on a parallel corpus of regular TED talk transcriptions and the ASR output generated for the TED talks that we described in the previous section. We ran a number of experiments to test the performance of such a model. Some of the results are listed in Table~\ref{tab:trans:data:subs:en-asr-results}.

\begin{table}[htbp]
    \begin{center}
    \begin{tabular}{llr}
        \hline
        Training data  & Model   & BLEU  \\
        \hline
        \tedasrtop{10} & \amun{} & 61.87 \\
        \tedasrtop{10} & \xf{}   & 61.91 \\
        \tedasrtop{50} & \amun{} & 61.82 \\
        \hline
    \end{tabular}
    \end{center}
    
    \caption{Translating English into ASR-like English using a model trained on \tedtrans{} and tested on the development test set with original ASR output as reference.}
    \label{tab:trans:data:subs:en-asr-results}
\end{table}

As expected, the BLEU scores are rather high as the target language is the same as the source language, and we only mutate certain parts of the incoming sentences. The results show that there is not such a dramatic difference between the different setups (with respect to the model architecture and the data selection) and that a plain attentional sequence-to-sequence model with recurrent layers (\amun{}) performs as well as a transformer model (\xf{}) in this case. This makes sense, as we do not expect many complex long-distance dependencies that influence translation quality in this task. Therefore, we opted for the \amun{} model trained on the top-$10$ ASR outputs, which we can decode efficiently in a distributed way on the CPU nodes of our computer cluster. With this we managed to successfully translate $99\%$ of the entire \subs{} collection from standard English into ASR-English. We refer to this set as~\subsasr{}.

We did a manual inspection on the result as well to see what the system actually learns to do. Most of the transformations are quite straightforward. The model learns to lowercase and to remove punctuation as our ASR output does not include it. However, it also does some other modifications that are more interesting from the viewpoint of an ASR module. While we do not have systematic evidence, Table~\ref{tab:trans:data:subs:en-asr-examples} shows a few selected examples that show interesting patterns. First of all, it learns to spell out numbers (see {\em ``2006''} in the first example). This is done consistently and quite accurately from what we have seen. Secondly, it replaces certain tokens with variants that resemble possible confusions that could come from a speech recognition system. The replacement of {\em ``E.U.''} with {\em ``you''} and {\em ``Stasi''} with {\em ``stars he''} in these examples are quite plausible and rather surprising for a model that is trained on textual examples only. However, to conclude that the model learns some kind of implicit acoustic model would be a bit far-fetched, even though we would like to investigate the capacity of such an approach further in the future.

\begin{table}[htbp]
    \begin{center}
    \begin{tabular}{rp{17.4em}}
           \hline
\textbf{Original} & \textbf{Because in the summer of 2006, the E.U.\linebreak{}Commission tabled a directive.} \\
       \asrref{}  &         because in the summer of two thousand and six the e u commission tabled directive  \\
       \asrout{}  &         because in the summer of two thousand and six you commission tabled a directive    \\
           \hline
    \end{tabular}
    \\[1em]
    \begin{tabular}{rp{17.6em}}
           \hline
\textbf{Original} & \textbf{Stasi was the secret police in East\linebreak{}Germany.}                           \\
       \asrref{}  &         what is the secret police in east germany                                          \\
       \asrout{}  &         stars he was the secret police in east\linebreak{}germany                          \\
           \hline
    \end{tabular}
    \end{center}
    
    \caption{Examples from the translations to ASR-like English. In the first column, \asrref{} refers to the top decoding hypothesis from the ASR model, while \asrout{} is the output of the model translating the output to an ASR-like format.}
    \label{tab:trans:data:subs:en-asr-examples}
\end{table}

In Section~\ref{sec:results}, we report on the effect of using synthetic ASR-like data on the translation pipeline.

\subsection{Recasing Experiments}
\label{sec:trans:recase}

Our first attempt at a post-processing experiment involved using case-insensitive translation models, and deferring case restoration to a separate process unconditioned by the source side that we would apply after translation. We used the Moses toolkit~\cite{moses} to train a recaser model on \tedtrans{}. Afterwards, we re-trained a translation model on \tedasrtop{10} and \subsasr{} after lowercasing the training and validation sets, re-translated the development test set with this model, and then used the recaser to restore cases in the lowercase translations that we obtained. As shown in Table~\ref{tab:trans:recase}, evaluating the translations produced through these additional steps yielded scores that were very similar to those obtained by the original case-sensitive translation models, and the result of this experiment was inconclusive.




\begin{table}[htbp]
    \begin{center}
    \begin{tabular}{lrr}
        \hline
        Training data                 & BLEU  & BLEU-\textsc{lc} \\
        \hline
        \tedasrtop{10}$+$\subsasr{}   & 19.79 & 20.43   \\
        \tedasrtop{10}$+$\subsasrlc{} & 19.73 & 20.91   \\
        \hline
    \end{tabular}
    \end{center}
    
    \caption{Case-sensitive models (\xf{}) versus lowercased models with subsequent recasing. Recasing causes a larger drop than the model gains from training on lowercased training data. BLEU-\textsc{lc} refers to case-insensitive BLEU scores.}
    \label{tab:trans:recase}
\end{table}

\subsection{Reranking Experiments}
\label{sec:trans:rerank}

In addition to using different subsets of the $n$-best lists output by the ASR model as additional training samples for the translation module, we also tried reranking alternatives using KenLM~\cite{kenlm}. We initially generated a tokenized and lowercased version of \tedtrans{} with all punctuation stripped, and then trained a language model on this set. We used this model to score and rerank samples in the $50$-best lists, and then generated a new top-$10$ subset from this reranked version. However, when we re-trained translation models from these alternative sets, we observed that the model trained on the top-$10$ subsets before reranking exhibited a significantly better translation performance. We suspect that this is because, while the language model is useful for assessing the surface similarity of the ASR outputs to the source-side references, it was not uncommon for it to assign higher scores to ASR outputs that are semantically inconsistent with the target-side references, causing the NMT module to produce erroneous translations.

Similarly, we experimented with another language model trained on the target side of \tedtrans{}, without the preprocessing. We intended this model to score and rerank outputs of the translation models, rather than the ASR module. To measure the effect of this language model, we fed the audio of our internal test set split through the ASR module, and produced $50$-best lists for each sample. Afterwards, we used the language model to score and rerank the alternative transcripts for each sample produced by translating this set, and then selected the highest-scoring output for each sample. As in the previous language model experiment, employing this additional procedure significantly crippled the performance of our translation models.


\section{Results}
\label{sec:results}

The results on development data reveal expected tendencies that we report below. First of all, as consistent with a lot of related literature, we can see a boost in performance when switching from a recurrent network model to the transformer model with multiple self-attention mechanisms. Table~\ref{tab:results:subs} shows a clear pattern of the superior performance of the transformer model that is also visible in additional runs that we do not list here. Secondly, we can see the importance of additional training data even if they come from slightly different domains. The vast amount of movie subtitles in \opensubs{} boosts the performance by about 3 absolute BLEU points. Note that the scores in Table~\ref{tab:results:subs} refer to models that do not use subtitles transformed into ASR-like English~(\subsasr{}) and which are not fine-tuned to TED talk translations.

\begin{table}[htbp]
    \begin{center}
    \begin{tabular}{llr}
    \hline
    Training data            & Model   & BLEU  \\
    \hline
    \tedasrtop{10}           & \amun{} & 16.28 \\
    \tedasrtop{10}$+$\subs{} & \amun{} & 19.93 \\
    \hline
    \tedasrtop{10}           & \xf{}   & 17.90 \\
    \tedasrtop{10}$+$\subs{} & \xf{}   & 20.44 \\
    \hline
    \end{tabular}
    \end{center}
    
    \caption{Model performance on the development test set when adding movie subtitles to the training data.}
    \label{tab:results:subs}
\end{table}


The effect of pre-processing by producing ASR-like English in the subtitle corpus is surprisingly negative. If we look at the scores in Table~\ref{tab:results:subs-asr}, we can see that the performance actually drops in all cases when considering only the untuned systems. We did not really expect that with the rather positive impression that we got from the manual inspection of the English-to-ASR translation discussed earlier. However, it is interesting to see the effect of fine-tuning. Fine-tuning here refers to a second training procedure that continues training with pure in-domain data (TED talks) after training the general model on the entire data set until convergence on validation data. Table~\ref{tab:results:subs-asr} shows an interesting effect that may explain the difficulties of the integration of the synthetic ASR data. The fine-tuned model actually outperforms the model trained on standard data, which is due to a substantial jump from untuned models to the tuned version. The difference between those models with standard data is, on the other hand, only minor.

\begin{table}[htbp]
    \begin{center}
    \begin{tabular}{lrr}
        \hline
                                      & \multicolumn{2}{c}{BLEU} \\
        Training data                 & Untuned  & Tuned         \\
        \hline
        \tedasrtop{10}$+$\subs{}      & 20.44    & 20.58         \\
        \tedasrtop{10}$+$\subsasr{}   & 19.79    & 20.80         \\
        \hline
    \end{tabular}
    \end{center}
    
    \caption{Training with original movie subtitles versus subtitles with English transformed into ASR-like English, before and after fine-tuning on \tedasrtop{10} as pure in-domain training data (\xf{}).}
    \label{tab:results:subs-asr}
\end{table}

The synthetic ASR data look more similar to the \tedasr{} data and, therefore, the model might get more confused between in-domain and out-of-domain data than it does for the model trained on the original subtitle data in connection with \tedasr{}. Fine-tuning to \tedasr{} brings the model back on track again and synthetic ASR data becomes modestly beneficial.


Also of note is the contrast between the evaluation scores we obtained in development and those from the official test set. The translations we submitted obtain the BLEU scores shown in Table~\ref{tab:results:test-set} on this year's test set.

\begin{table}[htbp]
    \begin{center}
    \begin{tabular}{lr}
        \hline
        Training data                 & BLEU     \\
        \hline
        \tedasrtop{10}                & 14.34    \\
        \tedasrtop{10}$+$\subs{}      & 16.45    \\
        \tedasrtop{10}$+$\subsasr{}   & 15.80    \\
        \hline
    \end{tabular}
    \end{center}
    
    \caption{BLEU scores from our final models (\xf{})---respectively, the 2nd contrastive, 1st contrastive, and primary submission---on this year's test set. The scores from the two models with \subs{} in their training data were obtained after fine-tuning on \tedasrtop{10}.}
    \label{tab:results:test-set}
\end{table}


\section{Conclusions}
\label{sec:concl}

Apart from employing well-established practices such as normalization and byte-pair encoding as well as the benefits of using the transformer architecture, the only substantial boost to translation performance came from our data selection for the NMT module. The NMT module of our best-performing system on this year's test set was trained on \tedasrtop{10} and the raw \subs{}, and later fine-tuned on \tedasrtop{10}.

Although we ran many experiments to improve various steps of our speech translation pipeline, their influence on translation performance has been marginal at best. The effects of training with different \tedasr{} subsets were hard to distinguish. While using \subsasr{} in training seemed to provide a modest improvement in development, this effect was not carried over to the final results on the test set. The later experiments with lowercasing and recasing had an ambiguous effect, and those with reranking had a noticeably negative outcome.

In future work, our aim is to further investigate what factors in a good speech translation model, and continue experimenting in relation to these on the NMT module. We will also try to improve our TDNN-HMM ASR module by replacing the n-grams with an RNNLM, and try see how our complete end-to-end speech-to-text translation model performs after having sufficient training time.



\section{Acknowledgements}
\label{sec:ack}

This work has been supported by the European Union's Horizon 2020 Research and Innovation Programme under Grant Agreement No 780069, and by the Academy of Finland in the project 313988.
In addition the Finnish IT Center for Science (CSC) provided computational resources.


\bibliography{iwslt2018}
\bibliographystyle{IEEEtran}

\end{document}